%% file: sample-sigconf-authordraft.tex
\documentclass[acmtog,authorversion]{acmart}

\AtBeginDocument{%
  }
    
\usepackage{multirow}
\usepackage{subcaption}
\usepackage{enumitem}

\setcopyright{none}
\renewcommand\footnotetextcopyrightpermission[1]{}

\usepackage{xcolor}
\definecolor{omniroamblue}{RGB}{0,114,178}

\begin{document}

\title{OmniRoam: World Wandering via Long-Horizon Panoramic Video Generation}

\author{Yuheng Liu}
\authornote{Work done during an intership at Adobe.}
\affiliation{
  \institution{University of California, Irvine}
  \country{USA}
}
\email{yuhenl22@uci.edu}

\author{Xin Lin}
\affiliation{
  \institution{University of California, San Diego}
  \country{USA}
}
\email{xil321@ucsd.edu}

\author{Xinke Li}
\authornote{Corresponding author.}
\affiliation{
  \institution{City University of Hong Kong}
  \country{Hong Kong, China}
}
\email{xinkeli@cityu.edu.hk}

\author{Baihan Yang}
\affiliation{
  \institution{University of California, San Diego}
  \country{USA}
}
\email{bay004@ucsd.edu}

\author{Chen Wang}
\authornotemark[1]
\affiliation{
  \institution{University of Pennsylvania}
  \country{USA}
}
\email{cw.chenwang@outlook.com}

\author{Kalyan Sunkavalli}
\affiliation{
  \institution{Adobe Research}
  \country{USA}
}
\email{sunkaval@adobe.com}

\author{Yannick Hold-Geoffroy}
\affiliation{
  \institution{Adobe Research}
  \country{USA}
}
\email{holdgeof@adobe.com}

\author{Hao Tan}
\affiliation{
  \institution{Adobe Research}
  \country{USA}
}
\email{hatan@adobe.com}

\author{Kai Zhang}
\affiliation{
  \institution{Adobe Research}
  \country{USA}
}
\email{kz298@cornell.edu}

\author{Xiaohui Xie}
\authornote{Equal contribution.}
\affiliation{
  \institution{University of California, Irvine}
  \country{USA}
}
\email{xhx@uci.edu}

\author{Zifan Shi}
\authornotemark[3]
\affiliation{
  \institution{Adobe Research}
  \country{UK}
}
\email{zshi@adobe.com}

\author{Yiwei Hu}
\authornotemark[3]
\affiliation{
  \institution{Adobe Research}
  \country{USA}
}
\email{yiwhu@adobe.com}

\renewcommand{\shortauthors}{Liu et al.}

\input{sections/0_abstract}



\begin{teaserfigure}
    \centering
  \includegraphics[width=0.94\textwidth]{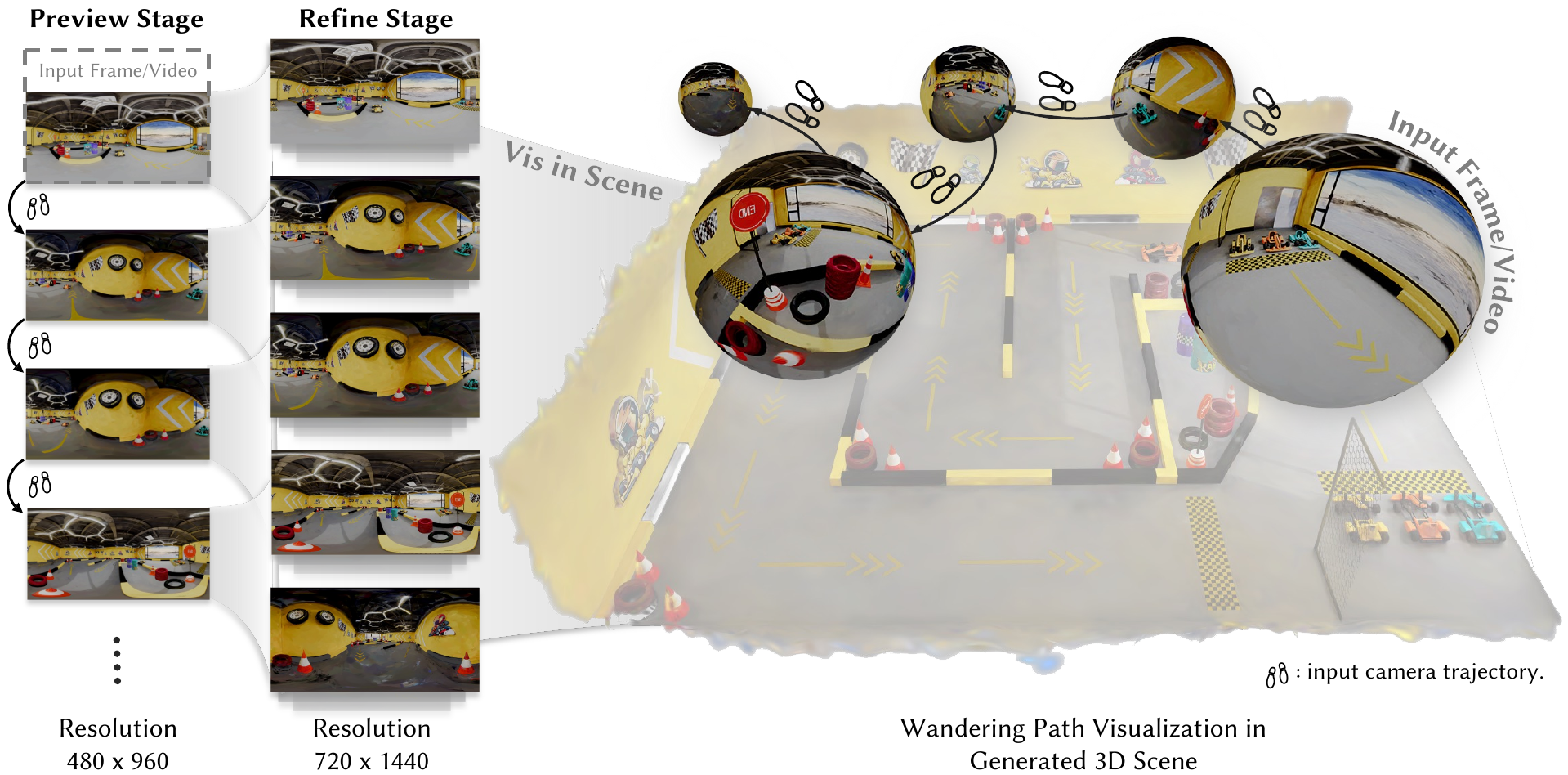}
  \vspace{-3mm}
  \caption{\textbf{Two-stage Long-horizon Panoramic Scene Wandering}. Starting from an input frame/video, \textbf{OmniRoam} first produces a coarse 480$\times$960 preview with accelerated playback speed to quickly traverse the scene along the user-specified camera trajectory (left). It then upscales this video to a high-quality 720$\times$1440 resolution (middle), jointly improving spatial details and temporal coherence, creating an immersive exploration experience. The resulting wandering path is visually represented as a series of viewpoints within the 3D scene (right).
  }
  \label{fig:teaser}
\end{teaserfigure}


\def\papername{OmniRoam}

\newcommand{\hyw}[1]{\textcolor{olive}{HYW: #1}}
\newcommand{\zf}[1]{\textcolor{cyan}{[Zifan: #1]}}
\newcommand{\yh}[1]{\textcolor{orange}{[Yuheng: #1]}}
\newcommand{\yhg}[1]{\textcolor{green}{[YHG: #1]}}
\maketitle

\input{sections/1_introduction}
\input{sections/2_related_work}

\input{sections/3_method}

\input{sections/4_experiments}
\input{sections/5_conclusion}
\input{sections/6_gallery}
\clearpage


\bibliographystyle{ACM-Reference-Format}
\bibliography{sample-base}


\end{document}

%% file: sections/0_abstract.tex
\begin{abstract}
Modeling scenes using video generation models has garnered growing research interest in recent years. However, most existing approaches rely on perspective video models that synthesize only limited observations of a scene, leading to issues of completeness and global consistency. 
We propose \textbf{\papername}, a controllable panoramic video generation framework that exploits the rich per-frame scene coverage and inherent long-term spatial and temporal consistency of panoramic representation, enabling long-horizon scene wandering. Our framework begins with a \textit{preview} stage, where a trajectory-controlled video generation model creates a quick overview of the scene from a given input image or video. Then, in the \textit{refine} stage, this video is temporally extended and spatially upsampled to produce long-range, high-resolution videos, thus enabling high-fidelity world wandering. 
To train our model, we introduce two panoramic video datasets that incorporate both synthetic and real-world captured videos. 
Experiments show that our framework consistently outperforms state-of-the-art methods in terms of visual quality, controllability, and long-term scene consistency, both qualitatively and quantitatively. We further showcase several extensions of this framework, including real-time video generation and 3D reconstruction. Code is available at {\color{omniroamblue}\url{https://github.com/yuhengliu02/OmniRoam}}.
\end{abstract}

%% file: sections/1_introduction.tex
\section{Introduction}
Generative models have recently achieved remarkable fidelity in creating visual content, such as images and videos~\cite{feng2025blobgen, zhang2025flashvideo, zhu2025generative}. However, generating coherent visual content at scale remains a fundamental challenge, particularly for scene modeling. Most existing scene generation methods~\cite{ho2020denoising, ho2022video, sohl2015deep, kim2024fifo, oh2024mevg} are grounded in perspective videos that reflect a human-centric viewpoint, which limits comprehensive modeling of the physical environment due to their limited field-of-view coverage. Recent autoregressive-based world models~\cite{blattmann2023stable, chen2024livephoto, guo2024sparsectrl, long2024videodrafter, RelicWorldModel2025} show promise for long-horizon video synthesis but often lead to accumulated errors and temporal drift during large camera movements~\cite{li2025stable, yu2025videossm}. These perspective videos present challenges, mainly because their view-limited nature complicates maintaining global scene consistency across space and time.

To address these limitations, we introduce panoramic vision as a foundation for large-scale scene generation. Unlike perspective views, panoramic representations inherently offer holistic scene coverage and global spatial memory~\cite{lin2025one,  lin2025depth, wang2025evoworld}, providing a robust basis for maintaining long-term consistency. In particular, we focus on the task of panoramic video wandering, in which the camera explores a scene along extended trajectories. This task allows each frame in a video to retain global context that would otherwise need to be accumulated over time, offering a natural and explicit mechanism for preserving spatial and temporal information in long-term video generation. Although recent work has begun to adapt existing generative architectures to panoramic data~\cite{liu2025dynamicscaler, zhang2025panodit, xing2025tip4gen, xie2025videopanda}, these efforts often treat panoramas as a direct extension of perspective imagery, without fully exploiting their representational advantages.

In this work, we present \textbf{\papername}, a controllable video generation framework for long-horizon panoramic video synthesis (Figure~\ref{fig:teaser}). 
We posit that effective scene modeling benefits from a global-to-local design: the global scene structure is first established by the panoramic representation, after which local details are progressively refined through increased spatial and temporal resolution, resulting in a scene representation that is both high-fidelity and globally coherent. Guided by this insight, we introduce a \textit{preview} stage that generates a coarse overview of the scene: Given a user-provided input image or video and a specified camera trajectory, our framework produces a panoramic video at mid-level spatial resolution with accelerated playback speed, enabling efficient navigation of large-scale scene layouts. In practice, thanks to the stochastic nature of generative models, users are able to efficiently generate multiple video variations and select their preferred scene configuration. Building upon the selected preview, a subsequent \textit{refine} stage operates on the generated panoramic video to synthesize the final high-resolution result at normal playback speed, supporting immersive, coherent long-horizon scene wandering. 

To train our models on a scalable dataset, we implement a panoramic data generation pipeline with a canonical coordinate system. We prepare both synthetic and real-world captured videos, providing accurate camera trajectories and high-fidelity visual quality. We also propose a novel metric, loop consistency, for evaluating long-term global consistency in scene generation, quantitatively demonstrating the benefits of panoramic representations.

Extensive experiments show our approach achieves better performance across multiple metrics, including visual quality, controllability, 
and long-term global consistency. 
We further demonstrate extensions of our framework, including a real-time preview mode 
and a 3D Gaussian Splatting (3DGS) reconstruction pipeline built from generated panoramic wandering videos. The primary contributions of our work are as follows:
\begin{itemize}[leftmargin=*, itemsep=0.2em, topsep=0.2em, parsep=0pt, partopsep=0pt]
\item We introduce a controllable framework for long-horizon panoramic video generation based on a global-to-local preview–refine design, enabling coherent and scalable scene wandering.
\item We demonstrate that panoramic representations and our design offer several advantages over perspective-based methods across multiple metrics, and we showcase extensions and applications using our framework.
\item We construct a scalable panoramic video data generation pipeline and will release a panoramic video dataset with camera trajectories to support large-scale scene modeling.
\item We propose a novel evaluation metric, loop consistency, to quantify long-term global spatial consistency in scene-level video generation, quantitatively demonstrating the advantages of panoramic representations.

\end{itemize}

%% file: sections/2_related_work.tex
\section{Related Work}

\noindent \textbf{Perspective Video Generation.} Video generation has rapidly advanced with the advent of diffusion models~\cite{ho2020denoising, ho2022video, sohl2015deep, peebles2023scalable, sora, xiao2025videoauteur, dalal2025one, kahatapitiya2025adaptive, feng2025blobgen, zhu2025generative, haji2025av}. Recent strong models like Hunyuan-Video \cite{kong2024hunyuanvideo}, CogVideoX \cite{yang2024cogvideox}, and Wan \cite{wan2025} achieve high-fidelity synthesis through architectural optimizations. However, maintaining long-term consistency remains difficult; standard and control-based methods often suffer from severe quality degradation or structural inconsistencies over time \cite{guo2024sparsectrl, long2024videodrafter}. While recent autoregressive approaches \cite{yin2025slow, huang2025self} attempt to address this, perspective-based methods remain inherently confined to a narrow field-of-view (FoV), limiting their ability to capture holistic scene contexts or maintain global spatial consistency during large-scale exploration.

\begin{figure*}[t]
  \centering
  \includegraphics[width = 0.98\linewidth]
  {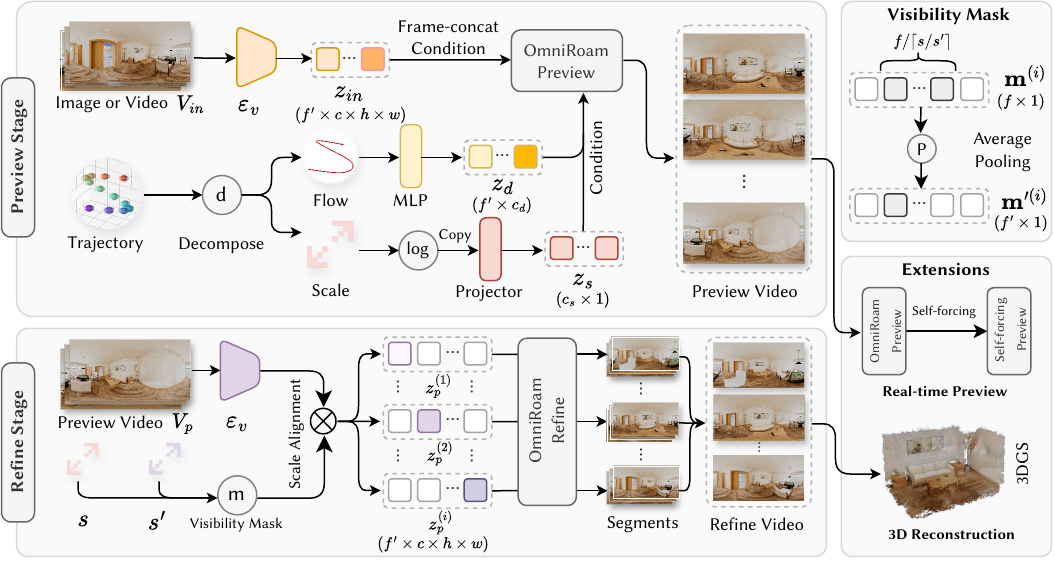}
  \vspace{-1em}
    \caption{\textbf{OmniRoam Pipeline} consists of (1) a \textit{preview} stage that takes an input image/video and a user-specified camera trajectory, and generates an accelerated, mid-resolution panoramic video via decomposed trajectory conditioning (flow + scale), and (2) a \textit{refine} stage that performs scale alignment and segment-wise diffusion to produce the final high-resolution long-horizon panoramic video. The framework further supports real-time autoregressive preview and downstream 3D reconstruction. Both models in \textit{preview} and \textit{refine} stage are finetuned from a pretrained Diffusion Transformers.
    }
    \label{fig:pipeline}
    \vspace{-1em}
\end{figure*}

\noindent \textbf{Panoramic Video Generation.} Panoramic vision offers holistic geometric cues to overcome the spatial limitations of perspective views. Early text-to-panorama frameworks generally fall into two categories: distortion-aware approaches, which adapt architectures to spherical geometry (e.g., DynamicScaler~\cite{liu2025dynamicscaler}, PanoDiT~\cite{zhang2025panodit}), and projection-driven methods that utilize multi-view or hybrid relationships for global coherence (e.g., TiP4GEN~\cite{xing2025tip4gen}, VideoPanda~\cite{xie2025videopanda}). 
More recently, WorldPrompter~\cite{zhang2025generating} leverages panoramic generation as an intermediate representation to synthesize globally coherent, 3D scenes for Gaussian Splatting reconstruction.

\noindent \textbf{Camera-controlled Video Diffusion Models.} Recent research injects explicit camera signals into video backbones to decouple dynamics from content, as seen in \cite{wang2024motionctrl, bai2025recammaster}. In the panoramic domain, 360DVD~\cite{wang2024360dvd} introduces spherical motion conditioning. More advanced frameworks, such as Matrix-3D \cite{yang2025matrix} and CamPVG \cite{ji2025campvg}, integrate trajectory guidance with sphere-aware mechanisms or 3D reconstruction to generate explorable worlds. 
However, these methods typically focus on short clips and suffer from long-horizon generation due to error accumulation. In contrast, ours employs a global-to-local preview-refine framework that exploits panoramic consistency to enable high-fidelity, scalable long-term scene wandering.

%% file: sections/3_method.tex
\section{\papername: Long Panoramic Video Wander}
\textbf{\papername} is a camera-controlled framework for long-horizon panoramic video synthesis based on a global-to-local generation strategy. As shown in Figure~\ref{fig:pipeline}, the framework consists of a \textit{preview} stage (Section~\ref{sec:method:preview}), which generates a mid-resolution panoramic video at accelerated playback speed conditioned on an input image and a camera trajectory, followed by a \textit{refine} stage (Section~\ref{sec:method:refine}) that produces the final high-quality panoramic video for immersive world wandering. We also describe our panoramic data generation pipeline (Section \ref{sec:method:dataset}), a novel long-term scene consistency metric (Section \ref{sec:method:metric}). In addition, we present extensions including real-time autoregressive preview and 3DGS reconstruction (Section \ref{sec:method:extension}).

\subsection{Trajectory-Controlled Video Preview}
\label{sec:method:preview}

The preview stage generates a panoramic video from user-provided images or videos and a specified camera trajectory, providing a rapid overview of the scene. We build on a pretrained video generation model~\cite{wan2025} and fine-tune it into a panoramic video generative model conditioned on visual inputs and explicit trajectory and motion-dynamics cues. A key innovation is to decompose camera motion control into two orthogonal factors: \emph{flow}, which specifies the normalized directions of movement, and \emph{scale}, which governs the global step size.

\noindent \textbf{Frame-Dimension Video Conditioning.} Following ReCamMaster~\cite{bai2025recammaster}, we adopt the frame-dimension conditioning strategy to ensure strict visual continuity. Let $V_{in} \in \mathbb{R}^{f \times  3 \times H \times W}$ denote the input video or image and $V_v \in \mathbb{R}^{f \times  3 \times H \times W}$ denote the target video to be generated, where $f$ is the temporal length. A 3D variational autoencoder $\mathcal{E}_v$ encodes them into latent representations:
\begin{equation}
    z_{in} = \mathcal{E}_v(V_{in}), \quad
    z_v = \mathcal{E}_v(V_v), \quad
    z_{in}, z_v \in \mathbb{R}^{f^{\prime} \times c \times h \times w}
    \label{eq:vae_encode}
\end{equation}
where $f^{\prime}, c, h, w$ denote the temporal length, channel dimension, and two spatial dimensions in latent space, respectively. Following ReCamMaster~\cite{bai2025recammaster} and Wan~\cite{wan2025}, the source and target latents are patchified and concatenated along the frame dimension.

\noindent \textbf{Decomposed Trajectory Conditioning.} 
 We decompose camera trajectory control into two orthogonal components: \emph{flow} and \emph{scale}. This decomposition enables intuitive speed adjustment for efficient scene exploration and provides finer-grained control: scale modulates motion globally across all frames, while flow specifies per-frame directional information. This separation also simplifies the subsequent refine stage, which only requires adjusting a single parameter (\textit{i.e.}, scale) without incorporating trajectory vectors. Furthermore, under two key assumptions—(a) \emph{uniform velocity}: the camera moves at almost constant speed along the trajectory, and (b) \emph{fixed orientation}: the camera orientation remains unchanged throughout the sequence, it is straightforward to decompose the trajectory into \emph{flow} and \emph{scale}. These assumptions are enforced during data curation, detailed in Section~\ref{sec:method:dataset}. 

\noindent\paragraph{Scale.} The scale $s \in \mathbb{R}^+$ is a global scalar representing the magnitude of camera displacement per timestep. We introduce it to quantify the playback speed and employ a $\text{log}$-space representation to handle a wide range of speeds. The scale embedding $z_{\text{s}}$ is computed via a projection:
\begin{equation}
    z_{\text{s}} = {\phi} (\log(s)),
    \label{eq:scale_embed}
\end{equation}
where $\phi \in \mathbb{R} \rightarrow \mathbb{R}^{c_s}$ is a learnable linear projector with output channel size $c_s$. This embedding is injected globally into the transformer blocks, uniformly modulating all temporal tokens.

\noindent\paragraph{Flow.} The flow $\mathcal{D} = \{\mathbf{d}_k\}_{k=1}^{f}$ is a sequence of $f$ unit vectors representing the normalized 3D direction of camera displacement at each timestep. Under the velocity decomposition, we have $\|\mathbf{d}_k\| = 1$ for all $k$. This directional information is encoded via a zero-initialized camera encoder $\mathcal{E}_c$:
\begin{equation}
    z_{d} = \mathcal{E}_c (\mathcal{D}),
    \label{eq:flow_embed}
\end{equation}
where $z_{d} \in \mathbb{R}^{f'\times {c_d}}$ is the embedding with channel size $c_d$, and encoder $\mathcal{E}_c$ is a fully connected layer inserted into each transformer block for frame-wise trajectory control.

\noindent \textbf{Training Objective.} Following the rectified flow framework in \citep{wan2025}, the forward process interpolates between data $z_v$ and noise $\epsilon \sim \mathcal{N}(0, \mathbf{I})$ as $z_t = t z_v + (1-t)\epsilon$.  During training, noise is added only to the target latent while keeping the source latent $z_{in}$ clean.  We concatenate the source and target latents frame-wise along the temporal dimension, where $z_{in}$ serves as a conditioning signal to ensure strict visual continuity. The loss is computed solely on the target portion:
\begin{equation}
    \mathcal{L}_{\text{preview}} = \mathbb{E}_{t, z_0, \epsilon} \left\| v_\Theta([z_{in}, {z}_t]_{\text{frame}}, t, \mathbf{c}) - (z_v - \epsilon) \right\|_2^2,
    \label{eq:preview_loss}
\end{equation}
where $v_\Theta$ is the predicted velocity field, $[\cdot]_{\text{frame}}$denotes frame-wise concatenation,   and $\mathbf{c} = \{ z_{d}, \tilde{z}_s\}$ denotes the conditioning set including flow embedding and scale embedding with temporal duplication.

\begin{figure}[t]
  \centering
  \includegraphics[width = 1\linewidth]
  {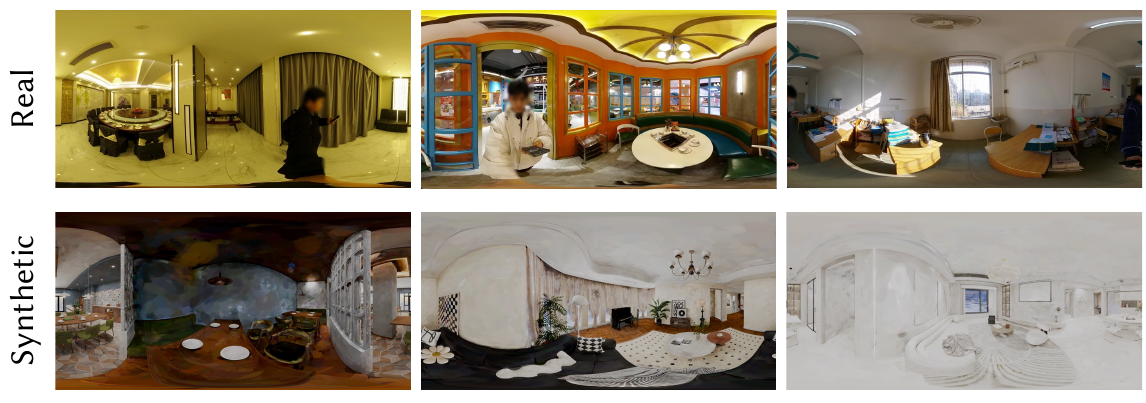}
    \caption{\textbf{Dataset Visualization.} We curate a hybrid training set containing real-world handheld panoramic videos captured across diverse scenes (top) and synthetic panoramic sequences rendered from 3DGS scenes with synthesized wide-ranging, physically feasible trajectories (bottom).
    }
    \label{fig:dataset}
\end{figure}

\subsection{Long-horizon Video Refinement}
\label{sec:method:refine}
While the preview stage generates camera-controlled videos at accelerated speed (\textit{i.e.}, a large scale  $s$), capturing extended trajectories requires temporal expansion and quality enhancement. Due to computational constraints, directly generating long, high-resolution videos in a single pass is infeasible. We introduce a second-stage diffusion model that performs segment-wise refinement through scale alignment and visibility mask conditioning. The refined $n$ segments are then concatenated to form the final long-horizon video $V = [V^{(1)}, V^{(2)}, \ldots, V^{(n)}]$.

\noindent \textbf{Scale Alignment and Visibility Mask Conditioning.} 
Given a preview video $V_p \in \mathbb{R}^{f \times 3 \times H \times W}$ generated at scale $s \in [1.1, 8.0]$ and a target scale $s'$ (\textit{e.g.}, $s' = 1.0$ for normal playback), we need to expand the temporal resolution by a factor of $s/s'$. We compute the number of segments required as  $n = \left\lceil \frac{s}{s'} \right\rceil$. For the $i$-th segment ($i = 1, \ldots, n$), a refined video sequence $V^{(i)}$ is synthesized with increased spatial and temporal resolution, conditioned on the selected preview frames. Specifically, we construct a binary visibility mask $\mathbf{m}^{(i)} \in \{0, 1\}^{f}$ to select which frames from $V_p$ serve as conditioning for generation:
\begin{equation}
    m_j^{(i)} = \begin{cases}
        1 & \text{if } j_0^{(i)} \leq j < j_0^{(i)} + w, \\
        0 & \text{otherwise},
    \end{cases}
    \label{eq:visibility_mask}
\end{equation}
where $w = \lceil (f-1) / n \rceil$ is the window size and $j_0^{(i)} = (i-1)w$ during inference. This mask selects corresponding preview frames as sparse conditions. 
During training, we randomly sample $j_0 \sim \mathcal{U}[0, f - w]$ for better generalization.

The preview is encoded as $z_p = \mathcal{E}_v(V_p) \in \mathbb{R}^{f' \times c \times h \times w}$, and the mask is downsampled to $\mathbf{m}'^{(i)} \in \mathbb{R}^{f'}$ via average pooling, as illustrated in Figure~\ref{fig:pipeline}. The masked latent conditioning is:
\begin{equation}
    \tilde{z}_p^{(i)} = z_p \odot \mathbf{m}'^{(i)},
    \label{eq:masked_latent}
\end{equation}
where $\mathbf{m}'^{(i)}$ is broadcasted across channel and spatial dimensions.

\noindent \textbf{Training Objective.} 
We concatenate the target latent $z'_v\in\mathbb{R}^{f'\times c \times h \times w}$ at scale $s'$ with $\tilde{z}_p^{(i)}$ along the frame dimension. Noise is added only to the target portion. The Rectified Flow objective is:
\begin{equation}
    \mathcal{L}_{\text{refine}} = \mathbb{E}_{t, j_0, z_v^{(i)}, \epsilon} \left\| v_\Phi([\tilde{z}_p^{(i)}, z'_t]_{\text{frame}}, t) - (z'_v - \epsilon) \right\|_2^2,
    \label{eq:refine_loss}
\end{equation}
where  $z'_t= tz'_v+ (1-t)\epsilon$, $z'_v$ is the target video segment at scale $s'$, and $v_\Phi$ is the velocity field predicted by the refine model.

\subsection{New Dataset and Benchmark} \label{sec:method:dataset}

The scale and precision of trajectory-aligned training data are crucial for high-fidelity and explorable panoramic video generation. We first establish a canonical panoramic coordinate system that decouples camera rotation from translation, tailoring the control space to spherical geometry.
Based on this protocol, we construct a hybrid dataset comprising scene-reality data derived from stabilized real-world footage and trajectory-precise data synthesized from 3DGS scenes with complex path planning (Figure~\ref{fig:dataset}). This design helps the model learn robust geometric priors from diverse, physically feasible motion patterns.

\noindent \textbf{Canonical Panoramic Coordinate System.} In an equirectangular projected (ERP) panorama, camera rotation corresponds to cyclic pixel shifts rather than geometric occlusion in perspective views. 
We thus design a rotation-invariant coordinate system that differs from standard 6-DoF approaches~\cite{yang2025matrix}.
By eliminating camera self-rotation (roll, pitch, yaw), we define trajectories strictly via translation $(x, y, z)$ relative to the ERP center ($\phi=0, \theta=0$). 
This formulation simplifies the generation space, ensuring that visual dynamics are driven exclusively by spatial translation relative to the anchor.

\noindent \textbf{Scene-Reality Real Data.}
We have constructed a large-scale dataset comprising approximately 2,000 handheld panoramic videos (5M frames) captured across diverse environments, including hotels, schools, and outdoor landscapes.
Following our rotation-invariant protocol, we gravity-align the footage and estimate camera trajectories using COLMAP~\cite{schonberger2016structure}.
We then filter out videos with abnormal scale to ensure that all data are approximately at the same scale.
Details are provided in the \textit{supplementary material}.

\noindent \textbf{Trajectory-Precise Synthetic Data.}
To construct a large-scale dataset with precise geometric supervision, we render 1,000 3D Gaussian Splatting (3DGS) scenes from InteriorGS~\cite{InteriorGS2025} and synthesize physically feasible trajectories via an automated pipeline.
We define the valid cruising area by filtering obstacles within the camera's vertical range (1.3m--1.5m) and generate candidate waypoints to cover 50\% of the free space.
We use a constant-speed trajectory for all frames to keep the scale consistent across videos.
Details are available in the \textit{supplementary material}.

\noindent \textbf{Metric for Long-term Scene Consistency.} 
\label{sec:method:metric}
Evaluating long-horizon spatial consistency is challenging, as standard metrics (\textit{e.g.}, FVD, FID) fail to capture global structural coherence over extended time. We therefore propose \textbf{Loop Consistency} ($\mathcal{C}_{\text{loop}}$), which measures whether a generated scene returns to its initial view after traversing a closed-loop camera trajectory. Given a generated sequence $V = \{I_i\}_{i=1}^{f}$ following a loop trajectory, we define:
\begin{align}
\mathcal{C}_{\text{loop}} &= \frac{1 - S_2}{1 - S_1}, \tag{8}\\
S_1 &= \frac{1}{P^2} \sum_{q=1}^{P} \sum_{p=1}^{P} \text{Sim}(I_p, I_{f-q+1}), \tag{9}\\
S_2 &= \frac{1}{P(f - 2P)} \sum_{p=1}^{P} \sum_{q=P+1}^{f-P} \text{Sim}(I_p, I_q). \tag{10}
\end{align}
where $\text{Sim}(\cdot, \cdot)$ denotes CLIP embedding cosine similarity. Here, $S_1$ (Loop Closure Score) measures visual similarity between the first and the last $P$ frames, while $S_2$ (Intermediate Score) captures its similarity to intermediate frames. A buffer of $P{=}5$ frames improves robustness against minor temporal misalignments. This metric rewards generations with high loop closure similarity ($S_1 \to 1$) while maintaining sufficient variation during exploration ($S_2 \to 0$).

\begin{table*}[t]
\centering
\caption{\textbf{Quantitative Comparison.} Our method consistently outperforms prior approaches~\cite{yang2025matrix, tan2024imagine360} across all evaluated metrics, achieving superior visual quality, stronger trajectory controllability, and higher loop consistency. FAED, SSIM, and LPIPS are evaluated on 81 frames, while loop consistency is reported on the full sequences.}
\vspace{-1em}
\resizebox{1\textwidth}{!}{
\begin{tabular}{ccc|ccc|ccc|c}
\toprule
\multirow{2}{*}{\textbf{Method}} 
& \multirow{2}{*}{\textbf{Resolution}} 
& \multirow{2}{*}{\begin{tabular}[c]{@{}c@{}}\textbf{No. of}\\ \textbf{Frames}\end{tabular}} 
& \multicolumn{3}{c|}{\textbf{Visual Quality}} 
& \multicolumn{3}{c|}{\textbf{Trajectory Controllability}} 
& \multirow{2}{*}{\textbf{Loop Consistency}} \\

& & 
& FAED$\downarrow$ 
& SSIM$\uparrow$ 
& LPIPS$\downarrow$ 
& PSNR'25$\uparrow$ 
& PSNR'55$\uparrow$ 
& PSNR'75$\uparrow$ 
&  \\ 
\midrule \midrule

Matrix-3D 
& 480p 
& 81 
& 8.64 
& 0.63 
& 0.37 
& 19.05 
& 17.35 
& 17.04 
& 1.38 \\

Imagine360 
& 480p 
& 81 
& 23.35 
& 0.33 
& 0.61 
& -- 
& -- 
& -- 
& -- \\

\textbf{Ours} 
& 480p 
& 81 
& \textbf{5.27} 
& \textbf{0.70} 
& \textbf{0.18} 
& \textbf{23.06} 
& \textbf{20.69} 
& \textbf{19.87} 
& \textbf{2.34} \\

\midrule 

Matrix-3D 
& 720p 
& 81 
& 9.76 
& \textbf{0.66} 
& 0.36 
& 19.06 
& 17.63 
& 17.12 
& 1.41 \\

\textbf{Ours} 
& 720p 
& 641 
& \textbf{5.07} 
& \textbf{0.66} 
& \textbf{0.33} 
& \textbf{19.75} 
& \textbf{18.59} 
& \textbf{18.24} 
& \textbf{1.96} \\

\bottomrule
\end{tabular}}
\label{tab:main_results}
\vspace{-1em}
\end{table*}

\begin{figure*}[t]
  \centering
  \includegraphics[width = 1\textwidth]
  {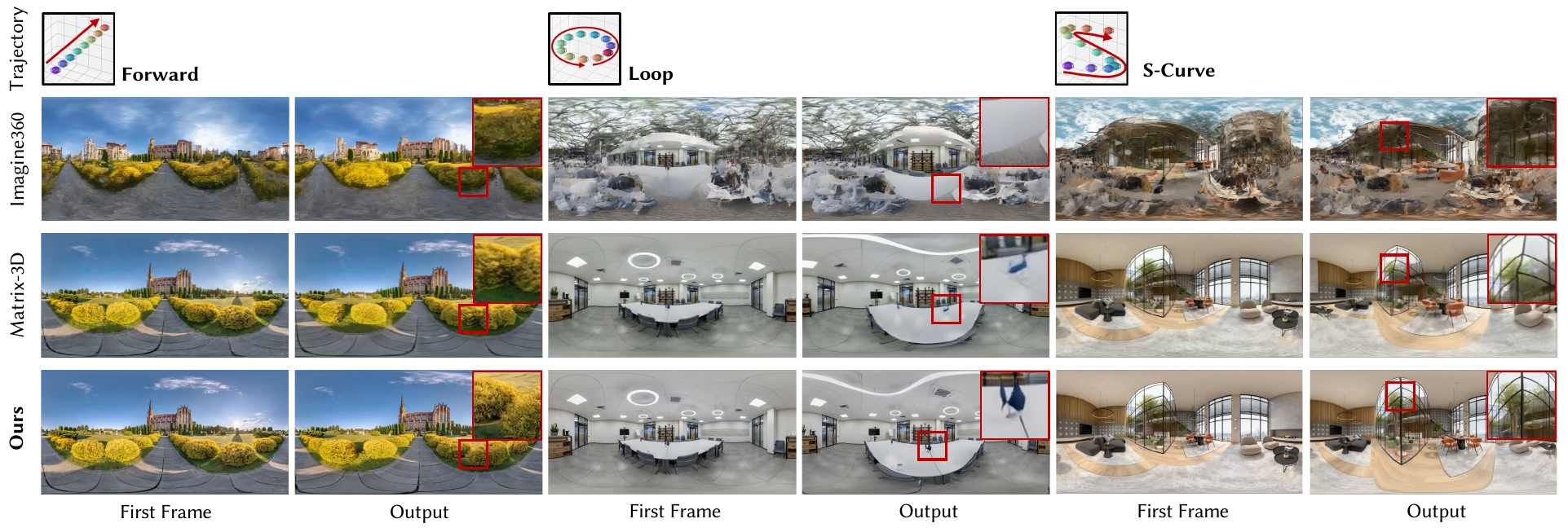}
  \vspace{-2em}
    \caption{\textbf{Qualitative Comparisons.} Our method generates panoramic videos that closely follow the input trajectory and preserve sharper details from the input images, while prior methods often produce blurry artifacts or semantically inconsistent content. Zoomed-in perspective crops are shown in the red inset.
    }
    \label{fig:main_comparison}
    \vspace{-1.2em}
\end{figure*}

%% file: sections/4_experiments.tex
\section{Experiments}

\subsection{Evaluation Protocols}
We evaluate our method from three aspects: \textit{visual quality}, \textit{trajectory controllability}, and \textit{loop consistency}. Evaluations are conducted at two output resolutions, 480p (Our Preview) and 720p (Our Refine). 
We define seven camera trajectories: \textit{forward}, \textit{backward}, \textit{left}, \textit{right}, \textit{s-curve}, \textit{loop}, and \textit{ground-truth (GT)}. None of these trajectories is directly included during training. For each trajectory and each method, we generate 24 videos using the test set. In qualitative visualizations presented in the paper, the input to the generation model is a single panoramic image collected from the Internet, in order to demonstrate generalization beyond the training data. Additional details are provided in the \textit{supplementary material}.

\noindent \textbf{Visual Quality.} We assess visual quality using Fr\'echet Auto-Encoder Distance (FAED)~\cite{zhang2024taming}, Structural Similarity Index (SSIM), and LPIPS.
FAED extends Fr\'echet Inception Distance by computing distances in an auto-encoder feature space, which reduces sensitivity to distortion artifacts introduced by equirectangular projection.
For SSIM and LPIPS, we generate videos conditioned on \textit{GT} trajectories and compare the synthesized results to the corresponding ground-truth videos frame-by-frame.

\noindent \textbf{Trajectory Controllability.} Directly comparing camera motion across different methods is unfair, as the effective motion scale may vary, and extracting camera poses from generated videos using structure-from-motion introduces additional noise.
To avoid these issues, we follow the evaluation protocol of CamPVG~\cite{ji2025campvg} and condition all methods on the same \textit{GT} trajectories.
We compute the average PSNR over three temporal windows: frames 20--25 (PSNR'25), 50--55 (PSNR'55), and 70--75 (PSNR'75).
For long video generation with $641$ frames, we additionally report PSNR over frames 610--615 (PSNR'615) and 630--635 (PSNR'635).
Higher PSNR values indicate more accurate adherence to the specified trajectory.

\noindent \textbf{Loop Consistency.}
We evaluate loop consistency using the metric introduced in Section~\ref{sec:method:metric}.
Specifically, we generate videos following loop trajectories and compute loop consistency scores over all frames.
A higher score indicates that the generated video returns to a visual state closer to the initial view after completing a full loop, reflecting stronger long-term spatial coherence.

\subsection{Experiment Settings}
\noindent \textbf{Model Training.}
We train two models for the preview and the refinement stages at resolutions of $480 \times 960$ (480p) and $720 \times 1440$ (720p), respectively.
Both models are finetuned on Wan2.1-1.3B~\cite{wan2025}.
The preview model generates 81-frame videos and supports panoramic images or videos, trajectory scale and flow as conditions.
We first train the model on real data with random temporal slicing for 120$k$ steps, using a batch size of 64 and a learning rate of $5 \times 10^{-4}$.
We then fine-tune the model on synthetic data for an additional 64$k$ steps with the same batch size and a reduced learning rate of $1 \times 10^{-4}$.
For the refine stage, the model takes temporally accelerated 480p video clips as input and reconstructs 720p videos at $1\times$ scale with 81 frames. Real and synthetic data are sampled with equal probability for training.
We use a batch size of 64 and a learning rate of $1 \times 10^{-4}$, and train the model for 120$k$ steps.

\noindent \textbf{Baselines.}
We compare our method against Matrix-3D \cite{yang2025matrix} and Imagine360 \cite{tan2024imagine360}.
Matrix-3D is finetuned on Wan2.1-14B, supports both 480p and 720p outputs, generates up to 81 frames, and allows panoramic image inputs with trajectory control.
Imagine360 converts perspective videos into panoramic videos and outputs 32-frame sequences at $512 \times 1024$.
For fair comparison, we downsample its outputs to 480p and temporally interpolate them to 81 frames.
We feed cropped perspective views from panoramas as input and primarily evaluate the quality of their generation.

\subsection{Results and Comparisons}

\noindent \textbf{Qualitative Results.}
Figure~\ref{fig:main_comparison} and Figure~\ref{fig:main_comparison_gallery} present qualitative comparisons with existing approaches using the same panoramic images collected from the Internet as input.
We visualize results under six different trajectories and include both indoor and outdoor scenes.
Our method consistently follows the specified trajectories while producing semantically coherent content with clearer object boundaries and more stable geometric structure.
In contrast, Imagine360 fails to generate meaningful panoramic content in most cases.
Matrix-3D is able to follow the input trajectory signals, but the generated videos often exhibit blurred appearance and noticeable geometric distortions, especially around object boundaries and structural elements.

We also develop an interactive system that enables users to generate videos interactively in both the preview and refinement stages; details are provided in the \textit{supplementary material}.

\begin{table*}[t!]
\centering
\caption{\textbf{Design Analysis.} We analyze key design choices, including video representation (ours: panoramic vs. perspective) and generation strategy (ours: global-to-local vs. direct autoregressive). We report FAED, SSIM, LPIPS, and loop consistency over the full video sequences. For long videos (641 frames), we additionally report the average PSNR over extended temporal windows, specifically frames 610–615 (PSNR'615) and 630–635 (PSNR'635).}
\vspace{-1em}
\resizebox{1\linewidth}{!}{
\begin{tabular}{cc|ccc|cccc|c}
\toprule
\multirow{2}{*}{\textbf{Method}} 
& \multirow{2}{*}{\begin{tabular}[c]{@{}c@{}}\textbf{No. of}\\ \textbf{Frames}\end{tabular}} 
& \multicolumn{3}{c|}{\textbf{Visual Quality}} 
& \multicolumn{4}{c|}{\textbf{Trajectory Controllability}} 
& \multirow{2}{*}{\textbf{Loop Consistency}} \\

& 
& FAED$\downarrow$ 
& SSIM$\uparrow$ 
& LPIPS$\downarrow$ 
& PSNR'25$\uparrow$ 
& PSNR'55$\uparrow$   
& PSNR'615$\uparrow$ 
& PSNR'635$\uparrow$ 
&  \\ 
\midrule \midrule

Perspective (Preview)
& 81 
& 16.90 
& 0.62 
& \textbf{0.44} 
& 17.72 
& 16.06 
& -- 
& -- 
& 1.70 \\

Perspective (Refine)
& 641 
& 15.48 
& \textbf{0.63} 
& 0.57 
& 16.60 
& 14.94 
& 13.76 
& 14.07 
& 1.42 \\

Autoregressive 
& 641 
& 16.04 
& 0.33 
& 0.64 
& 15.44 
& 14.84 
& 10.14 
& 10.11 
& 0.89 \\

\textbf{Ours} 
& 641 
& \textbf{7.70} 
& 0.58 
& \textbf{0.44} 
& \textbf{19.75} 
& \textbf{18.59} 
& \textbf{15.55} 
& \textbf{15.63} 
& \textbf{1.96} \\

\bottomrule
\end{tabular}}
\label{tab:ablation_results}
\vspace{-0.2em}
\end{table*}

\begin{figure*}[h!]
  \centering
  \includegraphics[width = 1\textwidth]
  {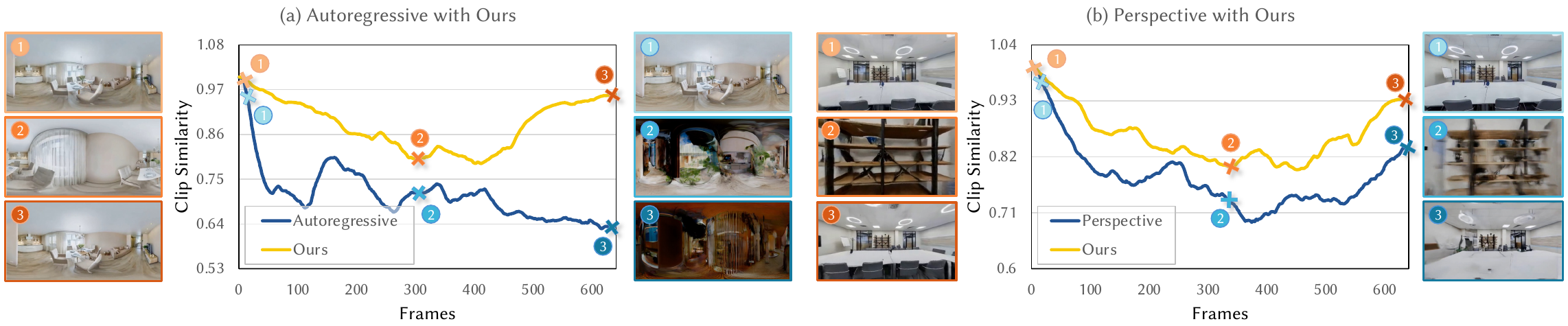}
  \vspace{-2.2em}
    \caption{\textbf{CLIP Similarity over Loop Trajectories.} We show temporal CLIP similarity to the first frame under loop trajectories for long-horizon generation (641 frames). Our method shows the expected trend: similarity decreases as the camera moves away and recovers as the trajectory closes the loop; whereas (a) the autoregressive variant exhibits a largely monotonic decline (drift from the initial view) and (b) the perspective-video variant shows weaker recovery with noticeable structural degradation. Representative frames are shown alongside the curves.
    }
    \label{fig:clip_similarity}
   \vspace{-1.2em}
\end{figure*}

\noindent \textbf{Quantitative Results.}
Table~\ref{tab:main_results} reports quantitative comparisons at 480p and 720p.
In terms of visual quality, our method outperforms baseline methods across all metrics, including FAED, SSIM, and LPIPS.
For trajectory controllability, we consistently outperform Matrix-3D across all three temporal windows in PSNR, indicating more accurate adherence to the specified camera trajectories. 
In terms of loop consistency, our method achieves higher scores than baselines, demonstrating stronger long-term spatial coherence.
We do not report trajectory controllability and loop consistency for Imagine360 because it does not support trajectory conditioning.
These results indicate that our model is able to return to visual states closer to the initial view after completing a loop, both for 81-frame sequences and for long sequences with 641 frames.

\noindent \textbf{Long-video Generation with Loop Trajectories.}
Beyond the loop consistency metric reported in Table~\ref{tab:main_results}, we further visualize the long-video generation results in Figure~\ref{fig:long-video_gallery}, where our method closely follows the loop trajectories over long distances and returns to the initial location with high visual consistency.
The last frames closely resemble the first frame of the sequence, demonstrating stable long-term behavior and effective temporal refinement.

\subsection{Design Analysis}
\noindent \textbf{Effect of Representation Choice.}
We compare panoramic and perspective representations by training perspective-video variants under identical settings with perspective data cropped from the panorama dataset.
As reported in Table~\ref{tab:ablation_results}, replacing panoramic videos with perspective videos consistently degrades visual quality, trajectory controllability, and loop consistency.
This degradation is observed in both short sequences generated by the preview stage (81 frames) and long sequences produced after refinement (641 frames).
These results indicate that the limited field of view in perspective videos makes it difficult to preserve global scene structure under large camera motion, even when using the same model design.

\noindent \textbf{Effect of Generation Strategy.}
To evaluate our global-to-local generation framework, we implement a straightforward autoregressive panoramic video generator that directly produces long video sequences by feeding each segment output into the next, 
trained under the same settings.
As shown in Table~\ref{tab:ablation_results}, the autoregressive variant performs worse than our full model across all metrics, with a particularly large gap in loop consistency, likely due to error accumulation across segments.
Our method achieves nearly twice the loop consistency score of the autoregressive baseline, indicating stronger long-term spatial coherence.

\noindent \textbf{Loop Consistency Analysis.}
Figure~\ref{fig:clip_similarity} provides a temporal analysis of consistent generation measured by clip similarity under loop trajectories.
For our method, similarity to the first frame decreases as the camera moves away from the starting position, and then gradually recovers as the loop closes, eventually approaching the similarity of the initial frame.
In contrast, the autoregressive variant (Figure~\ref{fig:clip_similarity}(a)) drifts monotonically, and the generated content at later frames deviates substantially from the initial panorama.
The perspective-video variant (Figure~\ref{fig:clip_similarity}(b)) also exhibits structural degradation.
Although the similarity curve shows a slight recovery near the end, the final frames differ significantly from the initial view, with noticeable loss of object geometry and scene details.
Together, these results demonstrate the necessity of our representation and design choices for stable long-horizon scene wandering.

\subsection{Extension and Application}
\label{sec:method:extension}
\noindent \textbf{Real-time Preview via Self-forcing.}
We extend our model to achieve real-time preview following self-forcing~\cite{huang2025self}, distilling the full model into a lightweight autoregressive previewer. 
Given a trained preview model (teacher), we match the real-time previewer's distribution (student) $p_{\hat{\theta}}$ to the teacher's distribution $p_\theta$ by minimizing $\mathbb{E}_{\hat{\mathbf{x}} \sim p_{\hat{\theta}},\mathbf{x} \sim p_{\theta}} \left[ D(\hat{\mathbf{x}}_{1:T}, \mathbf{x}_{1:T}) \right]$, 
where $\hat{\mathbf{x}}_{1:T}$ and $\mathbf{x}_{1:T}$ are sequences generated by the student and teacher, respectively, and $D(\cdot)$ is a distribution matching loss (\textit{e.g.}, discriminative loss or DMD score).
Our real-time previewer generates an 81-frame panoramic video in 7 seconds, substantially faster than the original preview-stage model ($\sim$ 5min) and significantly more efficient than Matrix-3D ($\sim$11min).
Figure~\ref{fig:previewer_and_3dgs}(a) shows that the real-time previewer preserves overall scene structure and produces plausible results. Additionally, Figure \ref{fig:self-forcing_gallery} demonstrates the results of generating 480p videos using the real-time model and subsequently enhancing them to 720p using our refine model. Please refer to the \textit{supplementary material} for more details.

\noindent \textbf{3D Scene Generation.}
We also showcase an application of our consistent long-horizon panoramic video generation framework for 3D scene generation and reconstruction. Specifically, given the generated long video comprising 641 frames, we uniformly extract 100 intermediate frames to represent the scene sequence. For each sampled panoramic frame, we crop five perspective views with a field of view of $120^\circ$ and a resolution of $512 \times 512$. These perspective crops are then used as input images for 3D Gaussian Splatting (3DGS)~\cite{3dgs} reconstruction.  Thanks to the long-range consistency of our generation, the reconstructed 3DGS scenes (Figure~\ref{fig:previewer_and_3dgs}(b)) have coherent structures across viewpoints, indicating that our panoramic videos offer sufficient view coverage and multi-view consistency to serve as a reasonable 3D representation. Please refer to the \textit{supplementary material} for more details.

%% file: sections/5_conclusion.tex
\section{Conclusion}
In this work, we present OmniRoam, a controllable panoramic video generation framework for long-horizon scene wandering. By using panoramic representations and a global-to-local preview–refine design, our method enables coherent exploration of large scenes with explicit camera control. We construct a panoramic video dataset with accurate trajectories and introduce a metric to evaluate long-term scene consistency. Experiments show that OmniRoam outperforms perspective-based methods in visual quality, controllability, and long-term coherence. We further demonstrate that the generated videos support real-time preview and downstream 3D scene reconstruction, highlighting the practicality of panoramic video generation for scene-level modeling.

%% file: sections/6_gallery.tex
\begin{figure*}[t]
  \centering
  \includegraphics[width = 0.98\textwidth]
  {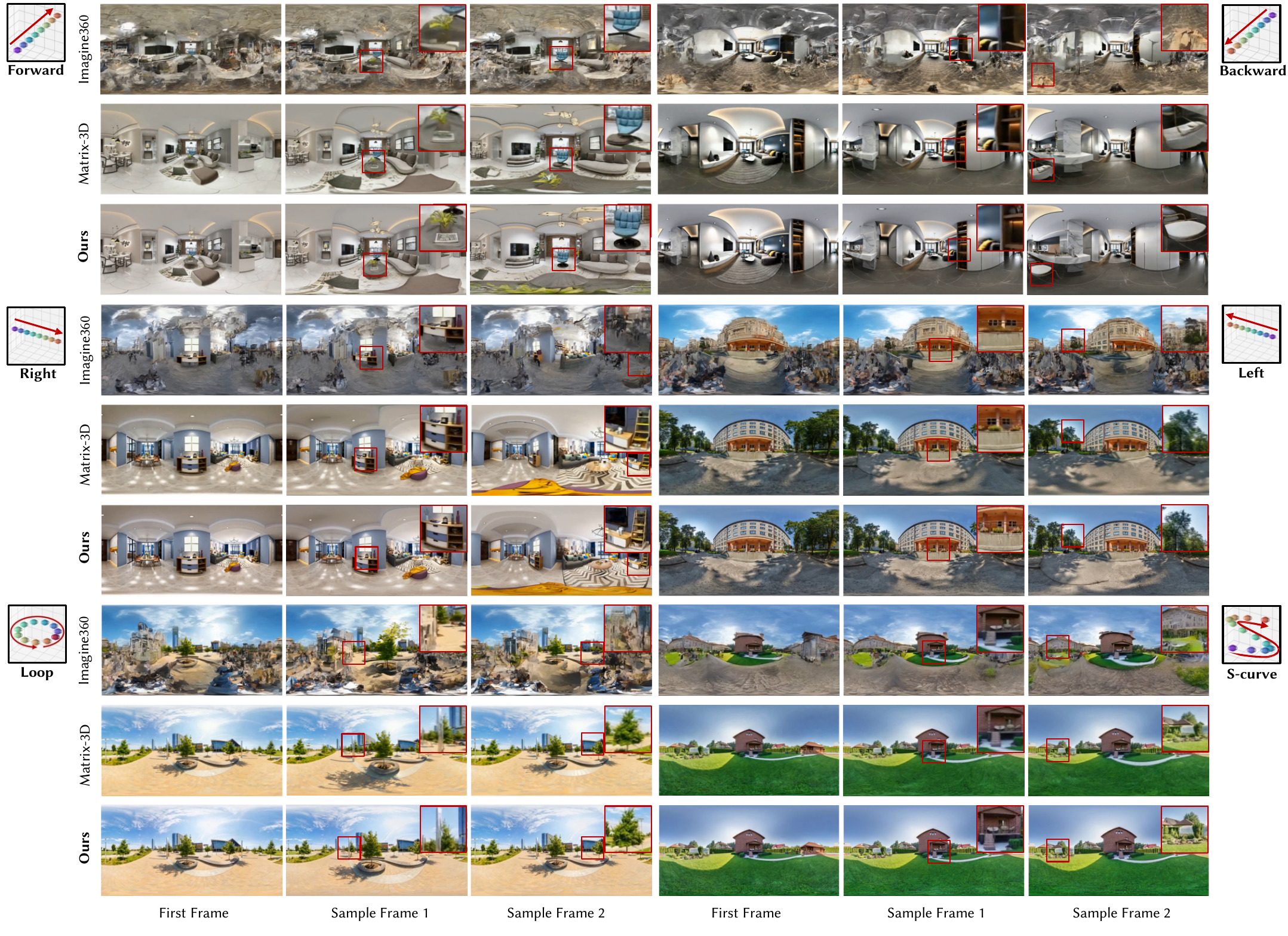}
    \caption{\textbf{Additional Qualitative Comparisons.} Our method maintains better geometric stability and finer textures during scene exploration.
    }
    \label{fig:main_comparison_gallery}
\end{figure*}

\begin{figure*}[t]
  \centering
  \includegraphics[width = 0.95\textwidth]
  {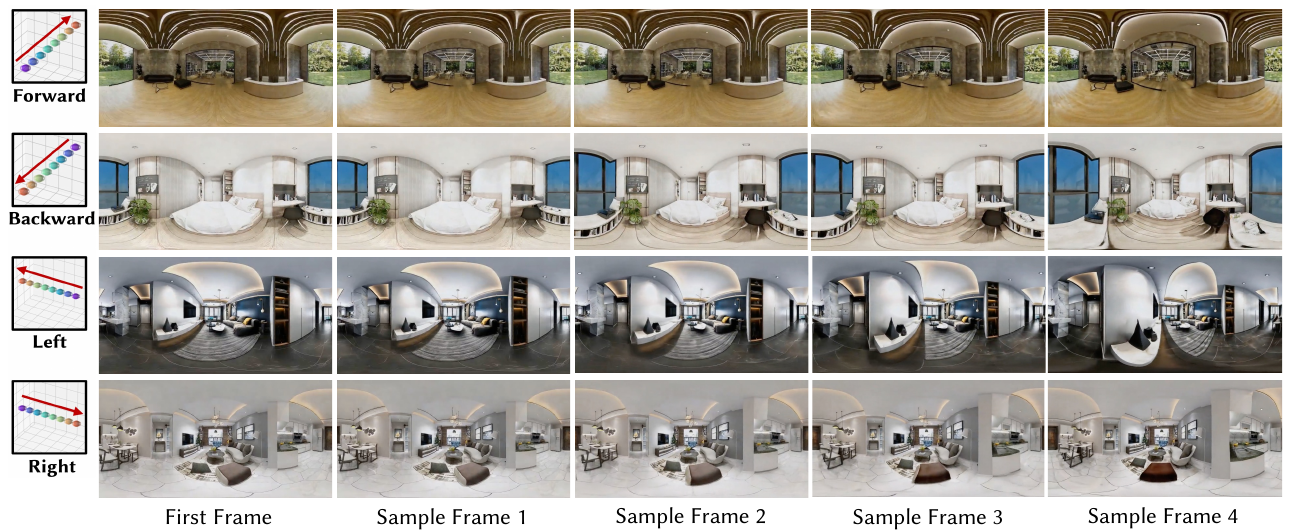}
    \caption{\textbf{Results from Real-Time Previewer-based Two-Stage Generation.} Videos are generated in real time at $480\text{p}$ and subsequently refined to $720\text{p}$.}

    \label{fig:self-forcing_gallery}
\end{figure*}

\begin{figure*}[t]

  \centering
  \includegraphics[width = 1\textwidth]
  {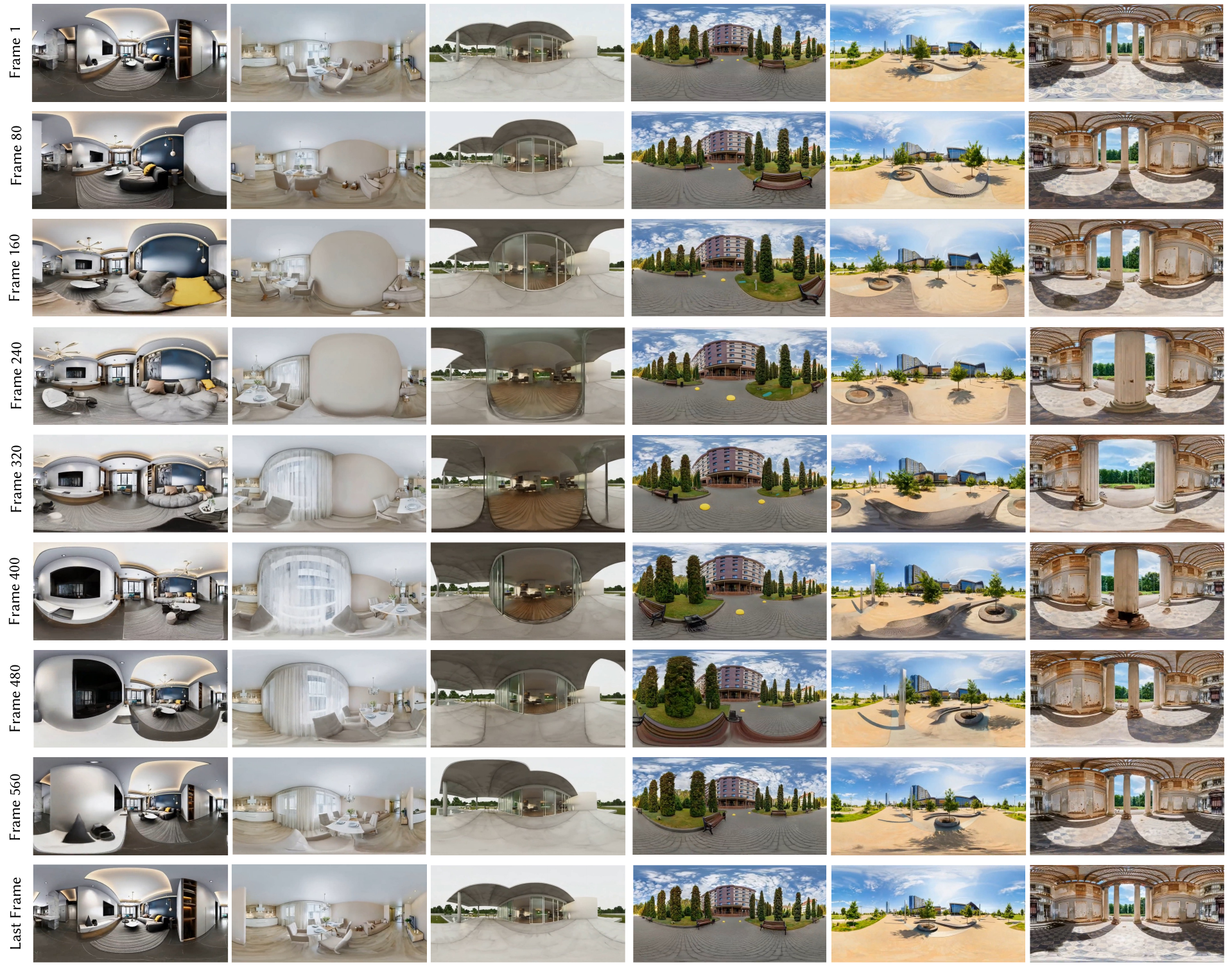}
   \vspace{-20pt}
    \caption{\textbf{Long Video Generation with Loop Trajectory.} We present 641-frame generated video sequences that traverse the scene in a closed loop, where the last frames closely resemble the initial frame, demonstrating strong long-term consistency.
    }
    \label{fig:long-video_gallery}
\end{figure*}

\begin{figure*}[!b]
  \centering

  \begin{subfigure}[t]{0.49\textwidth}
    \centering
    \includegraphics[width=\linewidth]{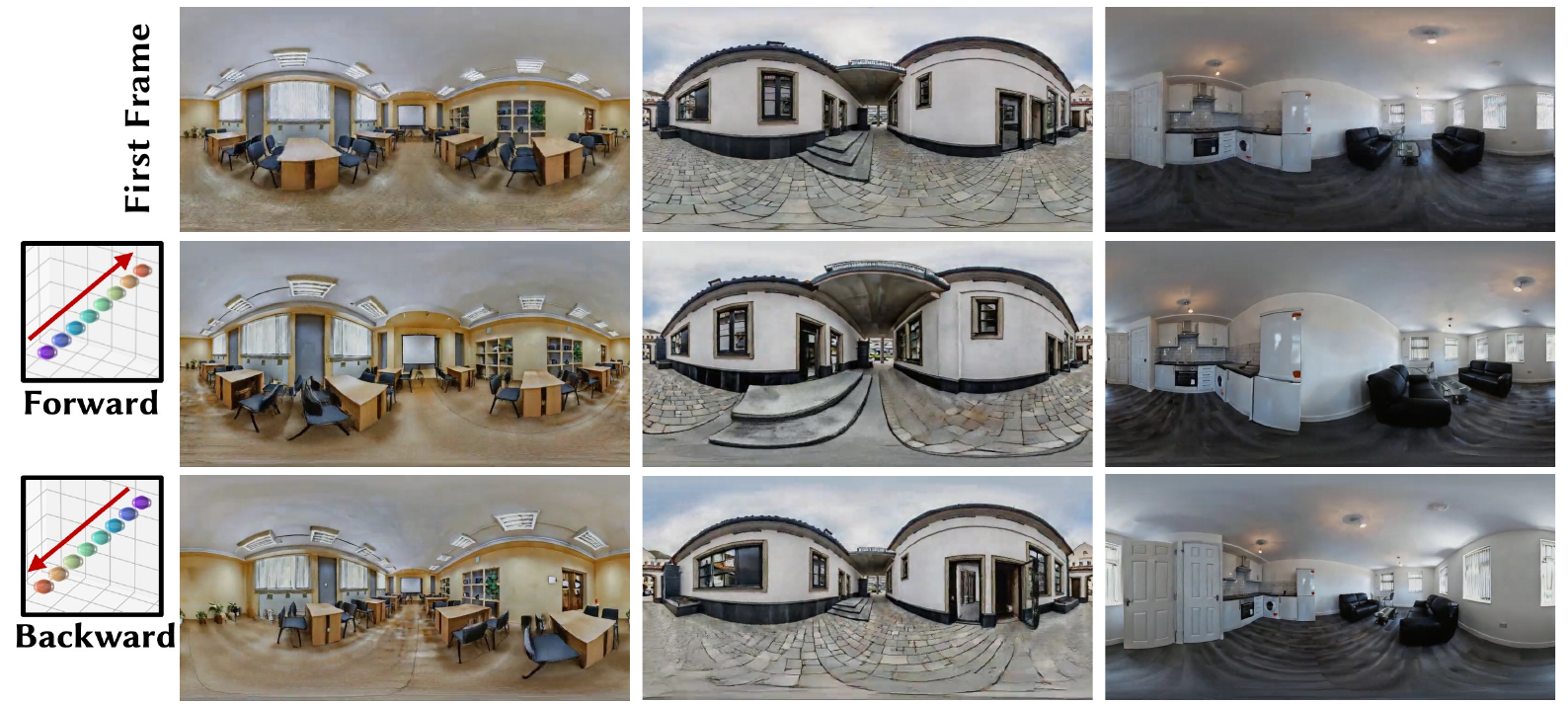}
    \caption{\textbf{Real-time Previewer.} The top row shows the input first frame, followed by generated frames along the specified camera trajectories.}
    \label{fig:previewer_a}
  \end{subfigure}
  \hfill
  \begin{subfigure}[t]{0.49\textwidth}
    \centering
    \includegraphics[width=\linewidth]{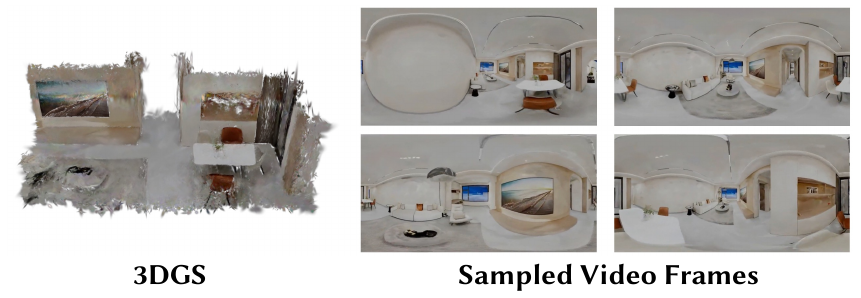}
    \caption{\textbf{3D Gaussian Splatting.} Left: reconstructed 3DGS; right: sampled panoramic video frames used for reconstruction.}
    \label{fig:previewer_b}
  \end{subfigure}

  \caption{\textbf{Visualization of Real-time Previewer and 3DGS.}
  (a) demonstrates trajectory-following panoramic wandering with our real-time previewer.
  (b) shows the reconstructed 3D Gaussian splatting results and corresponding panoramic samples.}
  \label{fig:previewer_and_3dgs}
\end{figure*}


